# Generative Models for Stochastic Processes Using Convolutional Neural Networks

**Fernando Fernandes Neto**
University of São Paulo - Brazil
fernando_fernandes_neto@usp.br

**Rodrigo de Losso da Silveira Bueno**
University of São Paulo - Brazil
delosso@usp.br

## Abstract

The present paper aims to demonstrate the usage of Convolutional Neural Networks as a generative model for stochastic processes, enabling researchers from a wide range of fields – such as quantitative finance and physics – to develop a general tool for forecasts and simulations without the need to identify/assume a specific system structure or estimate its parameters.

## 1 Introduction

It is widely known that procedures of identification, estimation and simulation of stochastic processes are somewhat already well established in mathematics, computing and related fields and sub-fields such as statistics, physics and econometrics. On the other hand, most of them rely on the fact that, to work properly, the observer must impose a system structure to estimate its respective parameters – see Hamilton [1994] and Hayashi [2000], and, very often, assume a probability distribution function.

On top of this specified and estimated system, the researcher can accomplish tasks such as forecasting and simulating the system's states.

In parallel to these strategies, it is possible to verify the revival of artificial neural networks, which have mostly been forgotten during the 1990s and mid-2000s because they were considered black-boxes without any intelligibility of its inner computations, as can be seen in Benitez *et. al* [1997] and in Kolman and Margaliot [2007].

One of the most important reasons for this revival is due to the success of Deep Neural Networks (neural networks with several layers), which have been successfully employed on tasks such as pattern recognition, classification and prediction, performing better than humans do. These tasks encompass a wide range of different problems such as computer vision problems – character recognition, object recognition and others; audio processing; and defeating world-class human players in complex computer games, Go and Chess. [see Silver *et. al*, 2017; Oshri and Khandwala, 2015].

A great part of these successes has been attributed to a new hybrid architecture called Convolutional Neural Networks, where convolutional filters are placed and stacked composing deep networks – enabling the filtering of desired multiscale/multidimensional features that enhance classification/forecasting capabilities – that can be interpreted and understood – plus a Softmax classifier/regressor, which basically works as a normalized multinomial logistic regressor – see Bishop [2006].

Given this huge success, the idea of the present paper is to adapt this specific kind of deep neural network, which has been successfully applied on the generation of raw audio waveforms as in Van den Oord *et. al* [2016-a], images as in Van den Oord *et. al* [2016-b], text [see Józefowicz *et. al*, 2016] and multivariate systems [see Borovykh *et. al*, 2017]; and show that this kind of neural network can also be used to work as a generative model on top of data retrieved from a wide set of known deterministic/stochastic data generation processes – from the simplest to the most complex processes, from damped oscillators to autoregressive conditional heteroskedastic (ARCH) and jump-diffusion models.

Avoiding the traditional identification and estimation procedures, a new approach is proposed here: estimate only the hyperparameters of a convolutional neural network, i.e. number of convolutional layers, discretization scheme (encoding) and dilations. Hence, we hope to demonstrate that data generation processes can be understood and simulated using a new statistical approach, without the need of assuming any hard-structural form or imposing any kind of restrictions. Moreover, as the data is encoded/decoded outside the neural network, by means of transforming a regression task into a classification task, no assumption about the distribution of the data generating process must be made. In addition to that, we demonstrate that the original data distribution can be recovered.

Potential applications are huge. Being able to simulate and predict stochastic processes properly is desired in a wide range of sub-fields within the scope of finance and economics, such as asset pricing, time series analysis and risk analysis.

To accomplish this goal, we have modified an existing Python/Tensorflow implementation of WaveNet as in Van den Oord *et. al* [2016-a], in order to read synthetic time series instead of raw audio files, avoiding any discussion about implementation strategies, focusing solely on the mathematical/statistical aspects of its usage. After that, we simulate each data generating process using a properly trained WaveNet, which their respective true parameters are known. Finally, using the R statistical package, the parameters of the simulated processes are estimated and analyzed.

That said, the paper is divided into these topics as follows:
- a brief description of the WaveNet architecture;
- description of the synthetic time series generated according to their respective generation processes;
- discussion of the hyperparameters chosen for each time series and a brief methodology of how to set up them;
- discussion of research findings;
- conclusions and propositions for future works.

## 2 A Brief Description of the WaveNet Architecture

The main idea of the original WaveNet paper [Van den Oord *et. al*, 2016-a] is to model the joint probability of a stochastic process $x = \{x_1, x_2, \ldots, x_T\}$ as a product of conditional probabilities

$$p(x_T) = \prod_{t=1}^{T} p(x_T \mid x_1, \ldots, x_{T-1})$$

In other words, the probability of $p(x_t)$ is conditioned to all previous observations.

To model the time series following this approach in terms of a Convolutional Neural Network, the WaveNet architecture consists of stacking what is called dilated causal convolutional layers, which consists of stacking structures as in Figure 1 and, as pointed before, a Softmax layer, which consists of a multinomial logistic classifier given by:

$$h_\theta(f) = \frac{1}{\sum_{j=1}^{K} \exp(\theta(j)^T f)} \begin{bmatrix} \theta(1)^T f \\ \theta(2)^T f \\ \ldots \\ \theta(K)^T f \end{bmatrix}$$

where $K$ denotes the number of different classes; $f$ denotes the features (independent variables) extracted in the previous layers; $\theta(j)^T$ denotes the weights of the features used to classify the output, which are filtered by means of convolution operations specified in Figure 1; and $h_\theta$ denotes the hypothesis of the output pertaining to a specific class – here, it is worth mentioning that, given this classification structure, the observed variables in the stochastic process must be encoded into a discrete variable with $K$ different classes, where:

$$h_\theta(x) = \begin{bmatrix} P(y = 1 \mid x, \theta) \\ P(y = 2 \mid x, \theta) \\ \ldots \\ P(y = K \mid x, \theta) \end{bmatrix}$$

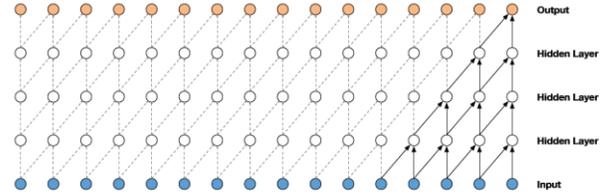

Figure 1: Dilation Causal Convolutional Layer. Source: Van den Oord *et. al*, 2016-a

In terms of the dilation causal convolutional layers, it is worth pointing out some important observations about the main features of these structures:

- stacked layers structures act as a generalization of Discrete Wavelet filters [see Borovykh *et. al*, 2017], given the fact that, basically, Discrete Wavelet transforms can be thought as a cascade of linear operations;
- stacking such features with non-linear operators can provide a general approximator due to the shift-invariance, as discussed in Bruna and Mallat [2013], Cheng *et. al* [2016] and Fernandes [2017], enabling the researcher to capture important non-linearities;
- this structure helps obtaining a very wide receptive field, which facilitates extracting long-range dependencies in conjunction with short memory (as can be seen in Figure 2).

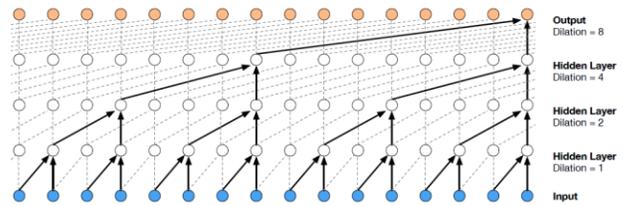

Figure 2: Multiscale features in Dilation Causal Convolutional Layers. Source: Van den Oord *et. al*, 2016-a

Keeping that in mind, these dilation convolutional layers are stacked and organized in a way that each feature is added to the previous layer features following a residual scheme, as follows in Figure 3, enabling deeper models and faster convergence, according to He *et. al* [2015].

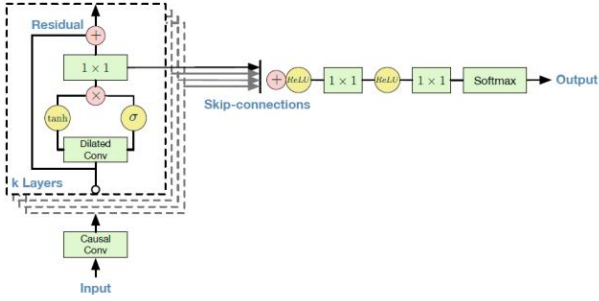

Figure 3: Overview of the Architecture. Source: Van den Oord *et.al*, 2016-a

It is also important to notice that there is a Gated Activation Function instead of Rectified Linear Activation Functions (ReLU), which are the most popular activation funciton in this kind of neural network, due to the fact that, as shown in Van den Oord *et. al* [2016-b], it outperforms the traditional approach. In this case, this Gated Activation Function is described by:

$$z = \tanh(W_{f,k} * x) \odot \sigma(W_{g,k} * x)$$

where $\odot$ denotes the element-wise multiplication operator; $*$ denotes the convolution operator; $x$ denotes the input; $W_{f,k}$ denotes a learnable convolutional filter at the $k$-th layer; and $W_{g,k}$ denotes a learnable convolutional gate.

Aiming to focus on the main subject of this paper, for further architecture details, one should read the original WaveNet implementation paper.

## 3 Description of the Synthetic Time Series

In the present paper, four deterministic data generation processes and five stochastic process were tested, in order to verify the generative capabilities of the WaveNet architecture. The chosen deterministic processes were:

- Harmonic Oscillator

$$\frac{d^2 y}{dt^2} + ay(t) = 0$$

- Damped Harmonic Oscillator (without any external forces)

$$\frac{d^2 y}{dt^2} + b\frac{dy}{dt} + ay(t) = 0$$

- Logistic Map with a chaotic choice of the parameter [see May, 1976]

$$x_{n+1} = r \cdot x_n \cdot (1 - x_n)$$

- Lorenz System, as simulated in Borovykh *et. al* [2017]

$$\begin{cases} \frac{dx}{dt} = \sigma(y - x) \\ \frac{dy}{dt} = x(\rho - z) - y \\ \frac{dz}{dt} = xy - \beta z \end{cases}$$

In these four deterministic cases, the Convolutional Neural Network should behave similar to a standard ordinary differential equation solver / difference equation solver. In the case of the stochastic processes, we have chosen the following processes:

- Standard diffusion process with mean-reversion

$$dX_t = \theta(\mu - X_t) + \sigma dW_t$$

- Jump-Diffusion Process (see Matsuda, 2004)

$$\frac{dX_t}{X_t} = (\alpha - \lambda k) \cdot dt + \sigma dW_t + (y_t - 1)dN_t$$

- Autoregressive Process of Order 1 (AR)

$$x_t = \phi \cdot x_{t-1} + \varepsilon_t + c$$

- Autoregressive-Moving Average Process of Order 1 (ARMA)

$$x_t = \phi \cdot x_{t-1} + \theta \cdot \varepsilon_{t-1} + \varepsilon_t + c$$

- Autoregressive Conditional Heteroskedastic Process of Order 1 (ARCH)

$$\begin{cases} x_t = \varepsilon_t h_t \\ h_t^2 = c + \phi_1 \cdot x_t^2 \end{cases}$$

In these cases, the Convolutional Neural Network should simulate a stochastic process compatible with the original one, as in a standard stochastic differential/difference equation simulator.

## 4 Description of the Methodology for the Choice of Hyperparameters

In order to setup the hyperparameters of the Convolutional Neural Network according to each synthetic time series, it was adopted a *forward* method, where we start with two layers and dilations up to the second order (i.e. dilations 1,2) and increased the dilations up to 256[th] order, which were the dilations in the original paper chosen to deal with eventual long-range dependencies found in text-to-speech applications.

In addition to that, it was generated a single time series (for each data generating process) with 12000 samples – 10000 samples were used to train the neural network and 2000 samples were used for back-testing purposes.

Whenever poor results were obtained, an additional convolutional layer was added, starting again with only dilations up to the second order.

Moreover, in this specific application, an 8-bit encoding was used to discretize the data into 256 classes; 256 skip-channels were used; with a filter width set equal to 2 – covering all the hyperparameters established in Van den Oord *et. al* [2016-a].

## 5 Simulations and Discussion of Research Findings

To test the capabilities of this architecture, we show first the results of the deterministic processes simulations and, afterwards, the results of the stochastic processes simulations. For visibility purposes, in the case of the Logistic Map (Figure 7), we show only the first 90 observations out-of-sample. It is also worth mentioning that, as we are modelling deterministic processes, in this case we use the *numpy.argsort()* method (present in the NumPy package for Python), to generate deterministic choices of the Softmax layer output, obeying the magnitude of the probabilities.

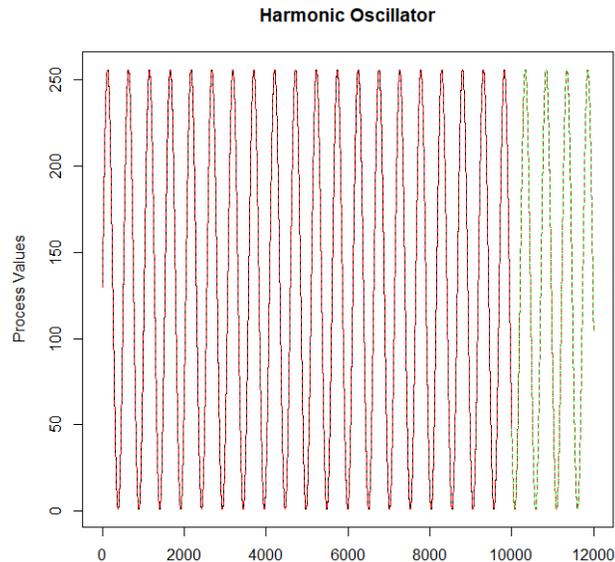

Figure 4: Simulation of a Deterministic Harmonic Oscillator (WaveNet with two layers with dilations up to 8)

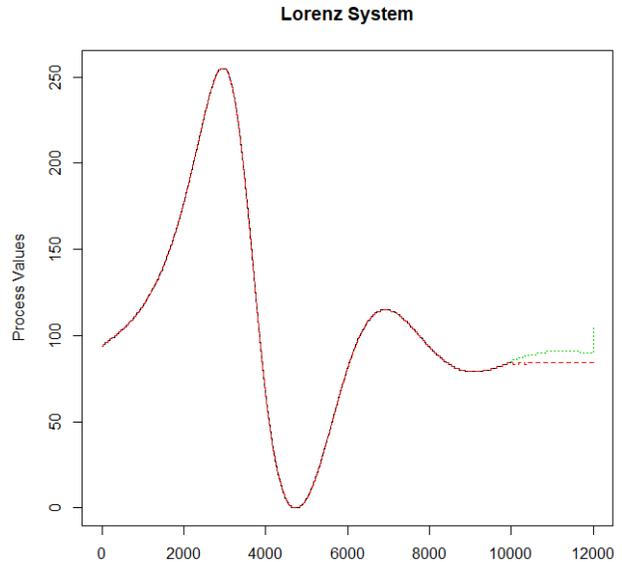

Figure 5: Simulation of a Deterministic Harmonic Oscillator (WaveNet with three layers with dilations up to 256)

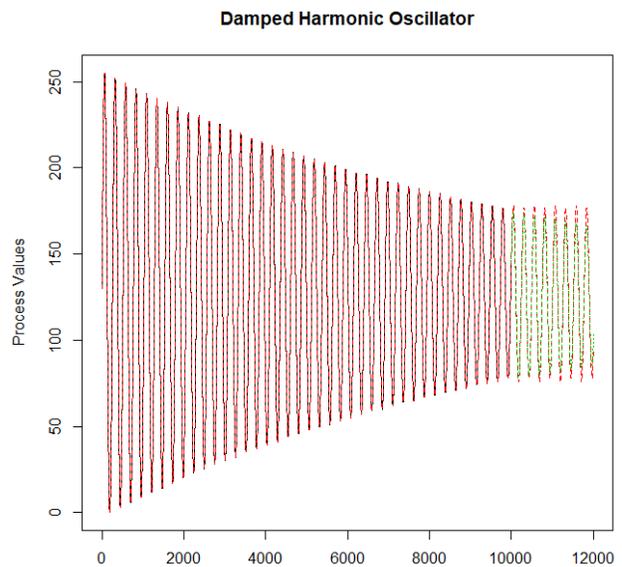

Figure 6: Simulation of a Deterministic Damped Harmonic Oscillator (WaveNet with nine layers with dilations up to 4)

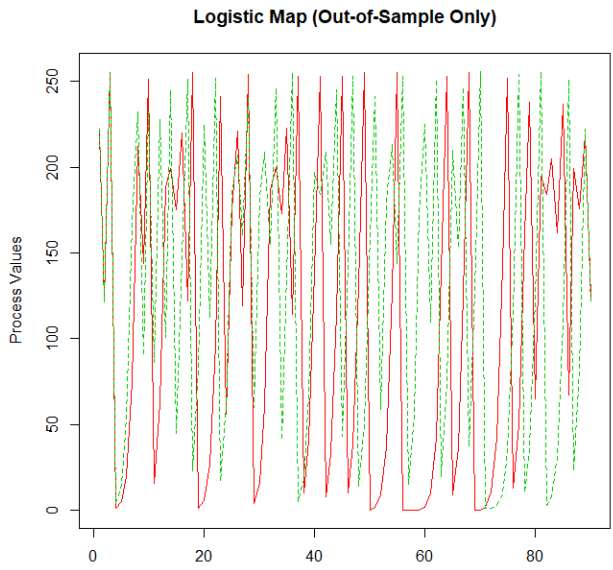

Figure 7: Simulation of a Deterministic Chaotic Logistic Map (WaveNet with five layers with dilations up to 2)

In the Figures 4 to 7, in all four graphics, dashed green lines represent observations used for back-testing purposes; the red lines represent predicted outputs by the model; and dashed black lines represent the original signal used to train the Neural Network.

Given that, it is possible to see that the WaveNet architecture is able to capture and fairly reproduce the main dynamics of the processes in Figures 4 and 5.

It is also interesting to notice that, in Figure 7, until the $10^{th}$ observation, the model is able to obtain precise out-of-sample forecasts of the data, while in the Figure 6, given the non-linear multivariable nature of the system, the best out-of-sample guess is an average of the past occurrences.

That said, we repeat the same experiment with the five proposed stochastic processes. However, instead of only plotting the time series (except for the jump diffusion process – plotted in Figure 8, which would require a lot of effort to be estimated, extrapolating the original scope of this present work), we also plot the distribution of the structural parameters of the simulated series.

Our hypothesis is supported by the fact that, if the structural parameters are compatible with the original ones – given the fact that we know the true data generating process parameters – the simulated process is compatible with the original one.

Hence, in Figure 8, we show first a Jump-Diffusion process with a negative drift, with a low probability of occurring a high intensity positive jump.

It is also worth noticing that in Figures 9 to 12, the median values are plotted in red and the true parameter values are plotted in blue. So, a direct comparison between the true known structural parameter values and the estimated one from the simulated series can be done.

To generate different realizations of the processes (represented by different colors), 100 simulations of each process were carried out, as in a Monte-Carlo approach, using the *numpy.random.choice()* method (present in the NumPy package for Python), to generate random choices that obey a given distribution, which is returned by the Softmax layer of WaveNet.

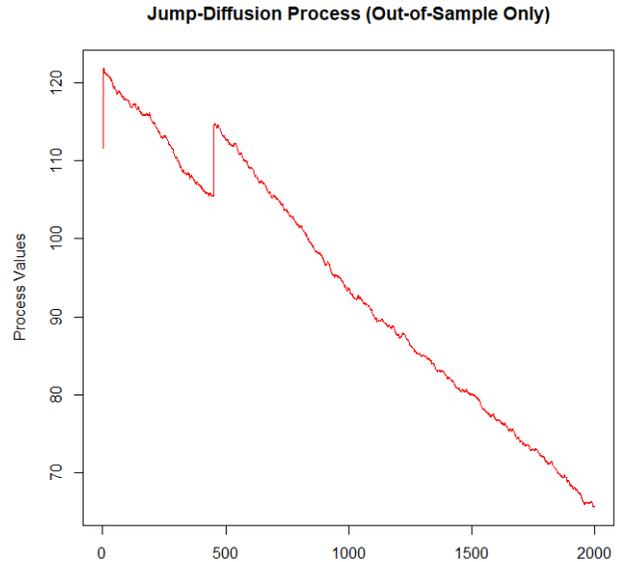

Figure 8: Simulation of a Jump-Diffusion Process (WaveNet with five layers with dilations up to 4)

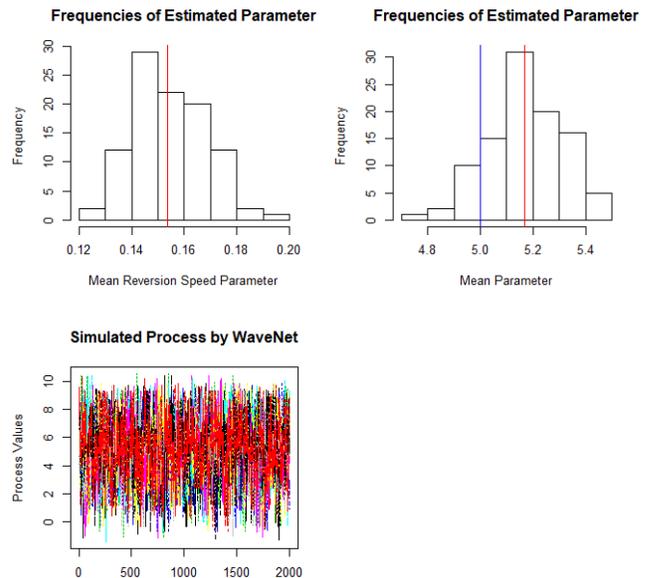

Figure 9 – Simulation and Inference of a Mean-Reverting Diffusion Process (True Mean Reversion Speed Parameter = 0.1 – blue line not shown)

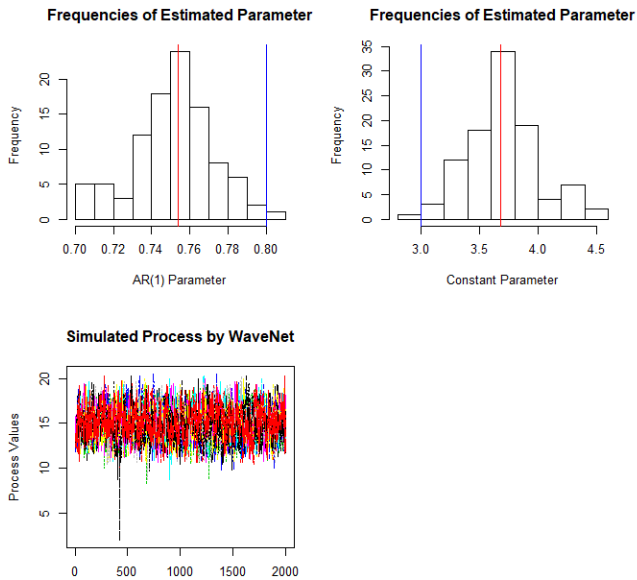

Figure 10: Simulation and Inference of an AR(1) Process (WaveNet with five layers with dilations up to 4)

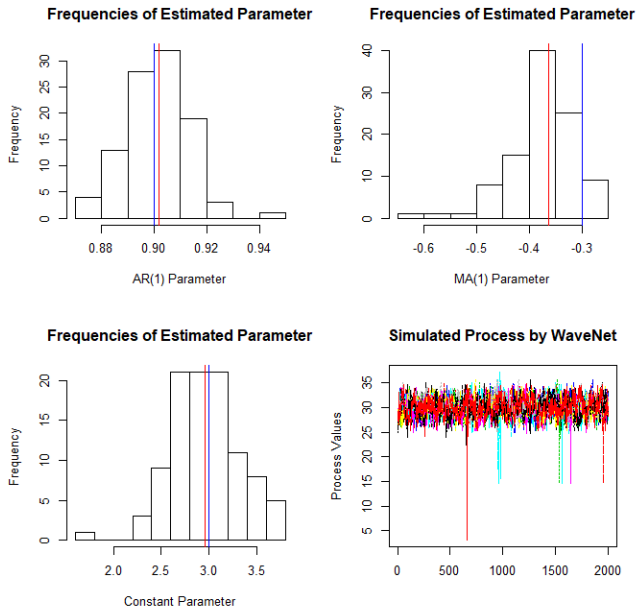

Figure 11: Simulation and Inference of an ARMA(1,1) Process (WaveNet with five layers with dilations up to 4)

In these three first linear models (Figures 9 to 11), it is possible to verify that the results are very reasonable, given the fact that the Networks have learnt from only one realization of each stochastic process (despite a large sample), and the parameters deviation from the true values are not large.

Also, it is important to state that, as all experiments used training sets where the first samples were equal to zero, it explains some of the valleys found in the simulations.

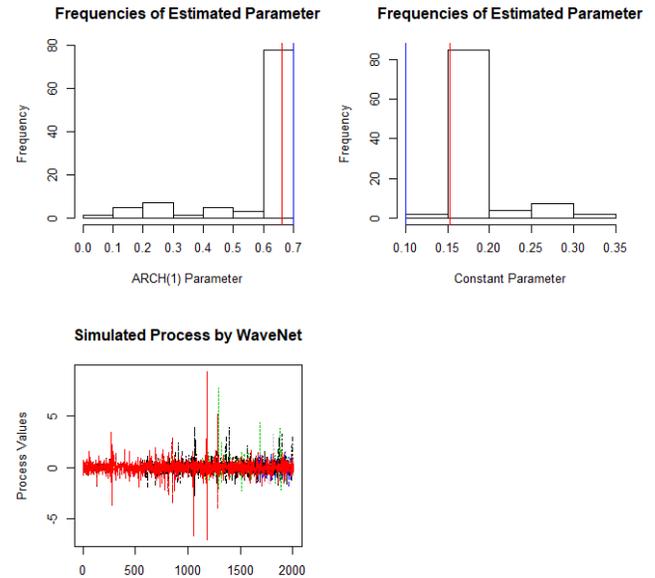

Figure 12: Simulation and Inference of an ARCH(1) Process (WaveNet with five layers with dilations up to 4)

In Figure 12, we can also verify that the structural characteristics of the data generation process are compatible with the true parameters.

## 6 Conclusions and Propositions for Future Works

In the present work, we aimed to establish a new simulation of data generation processes that avoid the traditional identification and estimation procedures, proposing here a new technique based on a Convolutional Neural Network – namely WaveNet architecture – where we estimate only its hyperparameters, in this case: number of convolutional layers, discretization scheme (encoding) and dilations.

To accomplish that, we have simulated different deterministic and stochastic processes, using an existing implementation of the WaveNet architecture code, adapted for this specific purpose, in conjunction with the R Statistical Package. On top of these different simulations, we show that the generated data is compatible with original data generation process, in a fair wide extent, being a potential attractive tool that can be employed in several different research areas.

As perspective for future works and research, we suggest following these experiments with more complex stochastic processes and, also, given the high computational cost of the training procedures (all of them were done using only one GPU in TensorFlow), it is important to develop an infor-mation criterion for such kind of models, in order to guide and speed-up the hyperparameters choice.

Moreover, extending this architecture for multivariate processes is a desirable path towards new interesting results.